\documentclass[11pt]{article}

\usepackage[final]{acl}
\usepackage{multirow}
\usepackage{times}
\usepackage{latexsym}
\usepackage{enumitem}
\usepackage{booktabs}
\usepackage{siunitx}
\usepackage[T1]{fontenc}

\usepackage[utf8]{inputenc}

\usepackage{microtype}

\usepackage{inconsolata}

\usepackage{graphicx}

\title{Beyond Trajectory Similarity: Evaluating Surgical Planning by Protocol Adherence}

\author{
  \textbf{
  Ruochen Li$^{1}$\thanks{Equal contribution.}
  \quad
  Kun Yuan$^{1,2}$\footnotemark[1]
  \quad
  Yufei Xia$^{3}$
  \quad
  Yue Zhou$^{1}$
  } \\
  \textbf{
  Qingyu Lu$^{4}$
  \quad
  Weihang Li$^{1}$
  \quad
  Youxiang Zhu$^{5}$\thanks{Joint supervision.}
  \quad
  Nassir Navab$^{1}$\footnotemark[2]}
   \\
  $^{1}$Technical University of Munich, Germany \\
  $^{2}$University of Strasbourg, France \\
  $^{3}$University of Glasgow, United Kingdom \\
  $^{4}$Nanyang Technological University, Singapore  \\
  $^{5}$University of Massachusetts Boston, USA
}

\begin{document}
\maketitle
\begin{abstract}
Surgical planning is commonly evaluated by trajectory similarity to a single observed execution, a paradigm borrowed from activity anticipation benchmarks such as Ego4D. This conflates two distinct concepts: \emph{task completion} and \emph{protocol adherence} and assumes similarity to a completed surgical trajectory implies a valid plan. In surgery, this assumption is flawed: the clinically relevant target is protocol adherence rather than mere completion, and the reference trajectory is only one observed execution that may itself deviate from the protocol.  We therefore introduce an evaluation framework grounded in expert-defined protocol rules that directly measures protocol adherence. Under this framework, similarity-based metrics systematically fail: they penalize protocol-adherent plans that deviate from the reference trajectory, while rewarding protocol-violating plans that remain superficially similar to it.  Evaluating Video-LLMs further reveals two separable failure modes: unreliable video-to-state grounding and under-constrained planning. Neither is resolved by model scaling alone, whereas structural knowledge injection provides consistent gains. Our results motivate a shift from trajectory similarity toward protocol adherence as the evaluation standard for surgical planning.
\end{abstract}

\begin{figure*}[t]
    \centering
    \includegraphics[width=0.8\linewidth]{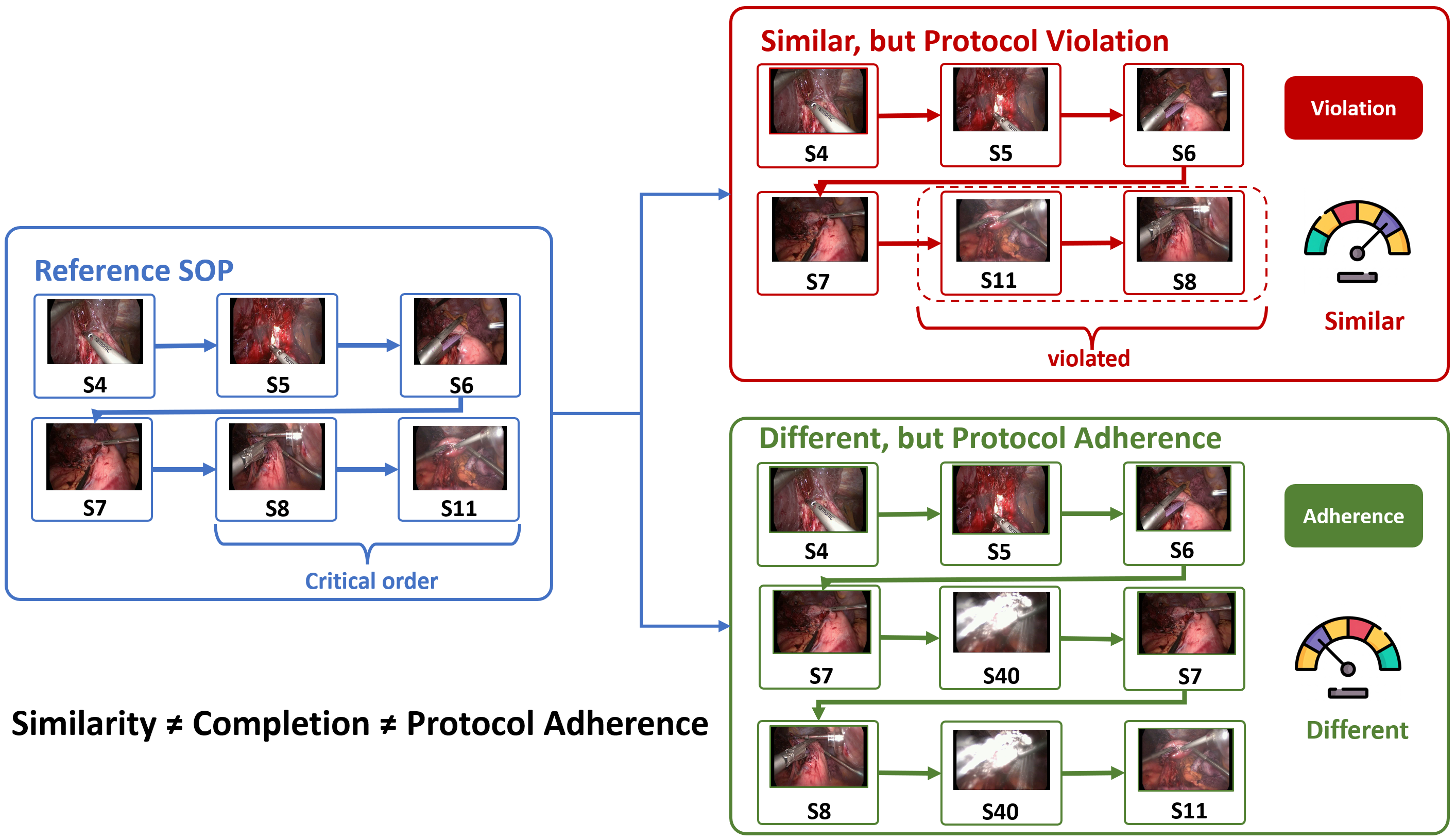}
    \caption{\textbf{Similarity $\neq$ Completion $\neq$ Protocol Adherence.}
    Both candidates complete the gastric pouch creation phase (P2), yet differ in protocol adherence.
    The upper candidate closely resembles the reference SOP but violates a critical ordering constraint (S8 appears before S11, reversing the required order).
    The lower candidate deviates from the reference yet fully adheres to the protocol by respecting all procedural dependencies.
    Trajectory similarity cannot distinguish these cases.
    Step abbreviations are listed in Appendix~\ref{appendix}.
}
    \label{fig:comparison}
\end{figure*}

\section{Introduction}

Vision-language models (VLMs) have become powerful foundation models capable of reasoning over visual content through natural language~\citep{zhang2025videollama}. In the video domain, they have progressed beyond low-level perception toward long-horizon temporal reasoning~\citep{Grauman2022Ego4D,Bai2025Qwen25VL}, enabling action recognition~\citep{fan2024vision}, anticipation~\citep{lin2024video}, and task-oriented planning~\citep{li2025optimus,Zhao2023AntGPT,li2025encouraging}, and supporting online assistance in everyday scenarios. As these capabilities mature, VLMs are increasingly considered for deployment in high-stakes settings, where errors carry significant consequences~\citep{Chen2025SurgPlanPP}. One such domain is surgery, where intelligent intra-operative surgical planning~\citep{Boels2025DARIL,Boels2025SWAG,Xu2025SurgicalPlanning} predicts future actions conditioned on the current procedural state and the goal, which is highly desirable but subject to strict requirements on safety and reliability.

Despite recent progress in surgical planning models, it remains unclear whether existing evaluation protocols reliably measure clinically valid planning behavior. Most prior work evaluates planning outputs by comparing predicted step or phase sequences to a single reference trajectory using surface sequence similarity metrics~\citep{ding2025understanding,Xu2025SAPBench} such as edit distance or relative order accuracy. This paradigm, common in activity anticipation benchmarks such as Ego4D~\citep{Grauman2022Ego4D}, treats trajectory similarity as a proxy for task completion. In surgery, this framing is flawed in two ways. First, similarity is a weak approximation of completion, where a plan can closely resemble the reference while omitting a critical step, or deviate from it while achieving all procedural goals~\citep{Lalys2014SPM,Biagini2025HieraSurg}. Second, completion alone is insufficient as an evaluation target. Surgical procedures are governed by pre-established clinical protocols whose requirements exist independently of any single observed execution. A surgeon may complete an operation while violating these requirements~\citep{wakabayashi2018tokyo,Lavanchy2025TechniqueImpact}. The clinically relevant evaluation target is therefore \emph{protocol adherence} (whether a plan satisfies the safety constraints defined by surgical consensus), rather than whether it merely completes the task or matches a reference trajectory.

To directly measure protocol adherence, we establish expert-defined surgical rules encoding hierarchical phase-step relations, required and permitted steps, procedural dependencies, and prohibited orderings. These rules operationalize protocol adherence as a conservative safety floor grounded in multicentric clinical standards. Building on this, we introduce a meta-evaluation benchmark grounded in MultiBypass140~\citep{Lavanchy2024MultiCentric}, comprising protocol-adherent procedural variations and systematically constructed protocol-violating plans with order and content errors. This benchmark enables us to test whether existing planning metrics align with protocol adherence rather than surface-level trajectory resemblance.

Using this benchmark, we show that widely used sequence similarity metrics are fundamentally misaligned with protocol adherence, and that LLM-as-a-judge baselines better  capture semantic omissions but fail to enforce strict procedural constraints.

We also use the expert-defined protocol rules to evaluate Video-LLMs under progressively informative planning settings, from end-to-end video planning to explicit state provision and knowledge injection. This reveals two separable failure modes: a perception bottleneck from unreliable video-to-state grounding, and an under-constrained reasoning space without explicit procedural structure. Structural knowledge injection yields consistent gains across model scales, whereas semantic descriptions alone remain insufficient.

These insights highlight the need for evaluation grounded in protocol adherence to support the development of reliable surgical planning systems. Contributions:
\begin{itemize}[topsep=2pt,itemsep=0pt,parsep=0pt,leftmargin=*]
\item We articulate why trajectory similarity fails for surgical planning evaluation, distinguishing task completion from protocol adherence.
\item We formalize protocol adherence via expert-defined surgical rules and construct a meta-evaluation benchmark with protocol-adherent variations and protocol-violating plans.
\item We benchmark sequence-similarity metrics and LLM-based judges, revealing systematic misalignment with protocol adherence.
\item We operationalize protocol adherence to evaluate Video-LLMs across progressively constrained planning settings, revealing perception and reasoning failure modes.
\end{itemize}

\section{From Trajectory Matching to Protocol Adherence in Surgical Planning}
\subsection{Trajectory Similarity Misaligns with Protocol Adherence}
Prior work treats trajectory similarity as a proxy for task \emph{completion}: if a predicted plan closely resembles the reference, it is assumed to capture the same goal-directed behavior~\cite{Xu2025SAPBench,Xu2025SurgicalPlanning,Boels2025SWAG}. 

As shown in Figure~\ref{fig:comparison}, this framing has two problems for surgical planning, first one is that \emph{completion is not the right target}. Surgery is governed by pre-established clinical protocols whose safety requirements exist independently of any observed execution~\citep{wakabayashi2020response,jin2022critical}. A surgeon can complete an operation while violating them. \emph{Completion} and \emph{protocol adherence} are distinct; for surgical planning, only the latter is clinically relevant.
The second problem is that similarity is a poor proxy even for completion: a plan can resemble the reference while omitting a critical step, or deviate from it while achieving all procedural goals. Since similarity fails to track completion, and completion is itself the wrong target. Trajectory similarity to a single reference cannot measure protocol adherence.

Clinical protocols exist precisely for safety and quality control: they encode the surgical community's consensus on how a procedure must be performed to minimize risk. Surgeries are expected to follow them. In practice, however, compliance is often incomplete. The Critical View of Safety (CVS) in laparoscopic cholecystectomy~\citep{strasberg2010rationale}, for example, is a widely endorsed safety standard requiring verification of specific anatomical conditions before the cystic duct is divided, yet studies report low implementation rates in real surgical practice~\citep{jin2022critical,manatakis2023critical}. Similar gaps between prescribed protocol and actual execution have been documented across surgical specialties~\citep{pucher2015sages,mascagni2020formalizing}.

\subsection{Observed Trajectories Do Not Represent the Protocol}
Using a recorded trajectory as the evaluation reference compounds both problems. A trajectory may itself be non-adherent, skipping safety steps or reflecting idiosyncratic surgeon habits, which makes the reference wrong. Even when a trajectory does adhere, it is only one sample from a broader space: surgical protocols permit local variation as long as safety constraints are satisfied, so different surgeons may follow different valid paths~\citep{wakabayashi2018tokyo,wakabayashi2020response}. In laparoscopic Roux-en-Y gastric bypass (RYGB)~\citep{Lavanchy2024MultiCentric, Lavanchy2025TechniqueImpact}, for instance, omentum division may precede gastric pouch creation without violating any protocol constraint. A single observed trajectory cannot represent the full space of adherent plans~\citep{twinanda2016endonet}, making it an unstable evaluation reference.

Together, these failures mean that a protocol-violating plan that stays close to a non-adherent trajectory can receive favorable scores while remaining clinically unsafe and passing evaluation entirely undetected.

\subsection{Expert-Defined Protocol Rules as Evaluation Ground Truth} \label{rules}
Protocol rules address both failure modes directly: they are derived from clinical consensus rather than observed trajectories, and they define a space of valid plans rather than a single reference. We therefore establish expert-defined protocol rules as the evaluation reference for protocol adherence. Rather than enumerating all valid plans, the rules specify safety-relevant constraints a plan must satisfy. A violation means the plan crosses a boundary defined by surgical consensus as clinically unsafe; satisfying all rules means the plan remains above the encoded safety floor, not that it is optimal or unique.

The rules are grounded in the MultiBypass140 annotation protocol~\citep{Lavanchy2024MultiCentric}, a multicentric collection of 140 laparoscopic RYGB procedures with surgeon-annotated hierarchical labels spanning 11 phases and 45 steps (each phase comprising a set of fine-grained surgical steps). Two experienced RYGB surgeons formalized the ordering requirements and phase-specific step exclusions already implicit in that protocol, without introducing independent clinical judgments. We define rules conservatively, covering only constraints directly derivable from the protocol and excluding center-specific or disputed preferences.

To clarify the components of rules, we analyze anastomosis test (Phase 5) as a case study.
\begin{itemize}[topsep=2pt,itemsep=0pt,parsep=0pt,leftmargin=*]
\item \textbf{Required Steps Set} defines mandatory steps to achieve the phase goal: S22 (gastric tube placement), S23 (jejunal clamping), S24 (dye injection), and S25 (visual assessment).
\item \textbf{Allowed Steps Set} enumerates additional steps that may plausibly occur without violating the phase objective, such as S3 (retractor placement), S39 (hemostasis), or S40 (irrigation aspiration).
\item \textbf{Procedural Dependencies} enforce critical ordering constraints among core steps, such as, if S24 occurs, the first occurrence of S23 must strictly precede it, and S25 must occur after S24.

\item \textbf{Prohibitive Orderings} further restricts when ancillary steps may appear. Steps such as S39 (hemostasis) or S40 (irrigation aspiration), if present, are permitted only after clamping has begun. Crucially, once the last S25 is completed, core test steps are prohibited from reappearing, explicitly marking completion of the phase.
\end{itemize}

A step sequence is labeled as \textbf{valid} if it satisfies all rules defined for the target phase; any rule violation renders it \textbf{invalid}.

\begin{table*}
  \centering
\resizebox{0.8\textwidth}{!}{%
  \setlength{\tabcolsep}{5pt}
  \sisetup{
    detect-weight=true,
    detect-family=true,
    table-number-alignment=center,
    table-space-text-post=\%
  }
  \begin{tabular}{c
                  S[table-format=2.1] S[table-format=2.1] S[table-format=2.1]
                  S[table-format=3.1] S[table-format=3.1] S[table-format=3.1] S[table-format=3.1] S[table-format=3.1] S[table-format=3.2]}
    \toprule
      & \multicolumn{3}{c}{\textbf{Traditional Metrics}}
      & \multicolumn{6}{c}{\textbf{LLM-based Judges}} \\
    \cmidrule(lr){2-4} \cmidrule(lr){5-10}
    \textbf{Subset} &
    \textbf{NED} &
    \textbf{JIS} &
    \textbf{ROA} &
    \textbf{Gemini 3 Pro} &
    \textbf{GPT 5.2} &
    \textbf{Claude 4.5 Sonnet} &
    \textbf{Hulu-Med 32B} &
    \textbf{Qwen3-32B} &
    \textbf{GPT-4o} \\
    \midrule
    \textbf{Valid} & 18.8 & 40.3 & 93.2 & 99.5 & 97.4 & 61.8 & 99.0 & 81.2 & 83.8 \\
    \midrule
    OE   & 87.3 & 46.5 & 11.3 & 11.3 & 18.3 & 80.3 & 8.5 & 35.2 & 38.0 \\
    CE   & 86.8 & 85.3 & 17.6 & 85.3 & 97.1 & 97.1 & 58.8 & 85.3 & 91.2 \\
    BE   & 96.7 & 85.0 & 20.0 & 98.3 & 100.0 & 100.0 & 75.0 & 95.0 & 93.3 \\
    \addlinespace[1pt]
    \textbf{Invalid} & 89.9 & 71.4 & 17.1 & 63.8 & 69.8 & 92.0 & 45.7 & 70.4 & 72.9 \\
    \bottomrule
  \end{tabular}
}
  \caption{
    Accuracy (\%) comparison of traditional similarity metrics and LLM-based judges
    across valid and erroneous subsets.
    NED: Normalized Edit Distance.
    JIS: Jaccard Index on Sequences.
    ROA: Relative Order Accuracy.
    OE: order error.
    CE: content error.
    BE: both error.
  }
  \label{tab:metric-comparison}
\end{table*}

\section{Can Existing Metrics Detect Protocol Violations?}
\subsection{Constructing a Protocol Adherence Evaluation Set}
We construct an evaluation dataset to test whether existing metrics accurately distinguish protocol-adherent from protocol-violating plans. It contains 191 valid sequences and 199 invalid sequences.

\textbf{Valid sequences} satisfy all rules and include protocol-adherent variations that differ from the MultiBypass140 defined Standard Operating Procedure (SOP). These cases test whether a metric incorrectly penalizes valid alternatives simply because they deviate from one recorded execution.

\textbf{Invalid sequences} violate at least one rule. We group them into three error types. Order errors violate procedural dependencies or prohibited orderings. Content errors include steps that are not appropriate for the corresponding phase. Both errors contain both ordering and content violations. This categorization allows us to test whether metrics separate valid from invalid plans and how they respond to different forms of protocol violation.

This setup turns metric evaluation into a binary discrimination problem. Given a valid or invalid plan under the expert-defined protocol rules, a metric should assign distinguishable scores to the two cases. A metric aligned with protocol adherence should therefore separate plans that satisfy the protocol rules from those that violate them, independently of their similarity to any particular observed trajectory.

\subsection{Evaluation Protocol}
We use a unified protocol to test whether existing planning metrics distinguish protocol-adherent step sequences from protocol-violating ones.

\textbf{Input.}
Each metric receives a candidate step sequence and a target surgical phase. For metrics that require a reference sequence, we use the SOP defined in MultiBypass140, which specifies the canonical step ordering for each phase.

\textbf{Output.}
Each metric produces a binary judgment indicating whether the candidate sequence is considered valid or invalid for completing the target phase. For continuous-valued metrics, scores are thresholded at 0.7 to obtain a binary decision.

\textbf{Comparison and Reporting.}
A metric aligned with protocol adherence should classify rule-conformant sequences as valid and rule-violating sequences as invalid. Performance is reported as binary classification accuracy, optionally stratified by sequence category (Valid, OE, CE, BE) to analyze metric sensitivity to different failure modes.

\subsection{Metrics under Comparison}
\textbf{Sequence Similarity Metrics.}
We evaluate a representative set of surface-level metrics, including Normalized Edit Distance (NED)~\citep{marzal2002computation}, Jaccard Index on Sequences (JIS)~\citep{broder1997resemblance}, and Relative Order Accuracy (ROA)~\cite{kendall1938new}, using their standard formulations. These metrics, together with their commonly used variants, have been widely adopted to measure sequence similarity and relative order agreement in prior surgical workflow analysis.

\textbf{LLM Judge.}
We evaluate several LLM-based judges that assess plan plausibility using injected phase-step relationships and descriptions, rather than the full explicit rules in Section~\ref{rules}. The full rule set is withheld because it defines the ground truth for protocol adherence. Providing it would expose the evaluation criterion and turn judging into rule execution. Each judge outputs both a binary validity decision and a textual explanation.

\subsection{Systematic Failure of Existing Metrics}

\textbf{Surface-level similarity metrics exhibit a systematic bias.}
NED and JIS achieve high accuracy on invalid sequences but perform poorly on valid plans (Table~\ref{tab:metric-comparison}), misclassifying the majority of clinically correct variations. This reveals the systematic bias described above: deviations from a single reference trajectory are penalized regardless of whether the sequence satisfies protocol constraints. Conversely, rule-violating sequences may receive favorable scores when they remain close to the reference. These metrics therefore conflate procedural variation with error because they evaluate behavioral imitation rather than protocol adherence.

\textbf{ROA is permissive but unsafe.}
ROA achieves high accuracy on valid sequences but fails catastrophically on order errors (Table~\ref{tab:metric-comparison}). By measuring only relative pairwise order, it overlooks repetitions and critical misplacements that render a procedure infeasible, thereby rewarding sequences that violate essential temporal and causal constraints.

\textbf{LLM-based Judges: Semantics over Structure.}
LLM-based evaluators perform well on content errors (Table~\ref{tab:metric-comparison}), indicating strong semantic understanding of phase goals and step relevance. They utilize internal medical knowledge to recognize that P2 cannot be completed if the \emph{stapling} step is missing.
However, most models struggle with order errors, frequently approving sequences that violate critical procedural dependencies. For phase P5, S24 is placed before S23, which would allow the dye to escape downstream before the test segment is sealed, thereby invalidating the anastomotic leak test; nevertheless, ChatGPT judged the sequence as correct, reasoning that all required steps for the test were present. This failure highlights a systematic tendency of LLMs to prioritize semantic completeness over strict procedural ordering, leading to errors when correct execution depends on temporal or causal constraints rather than the mere presence of steps.
Moreover, different models exhibit distinct biases. Hulu-Med tends to over-accept plausibly complete plans, while Claude over-reject them. Yet none consistently enforces dependency-level correctness.

Overall, existing metrics approximate protocol adherence through incomplete proxies.
These proxies systematically produce both false negatives, where protocol-adherent variants are penalized, and false positives, where safety-violating plans are rewarded.
Evaluating surgical planning therefore requires explicit procedural constraints rather than behavioral or semantic approximations.

\begin{figure*}
    \centering
    \includegraphics[width=0.95\linewidth]{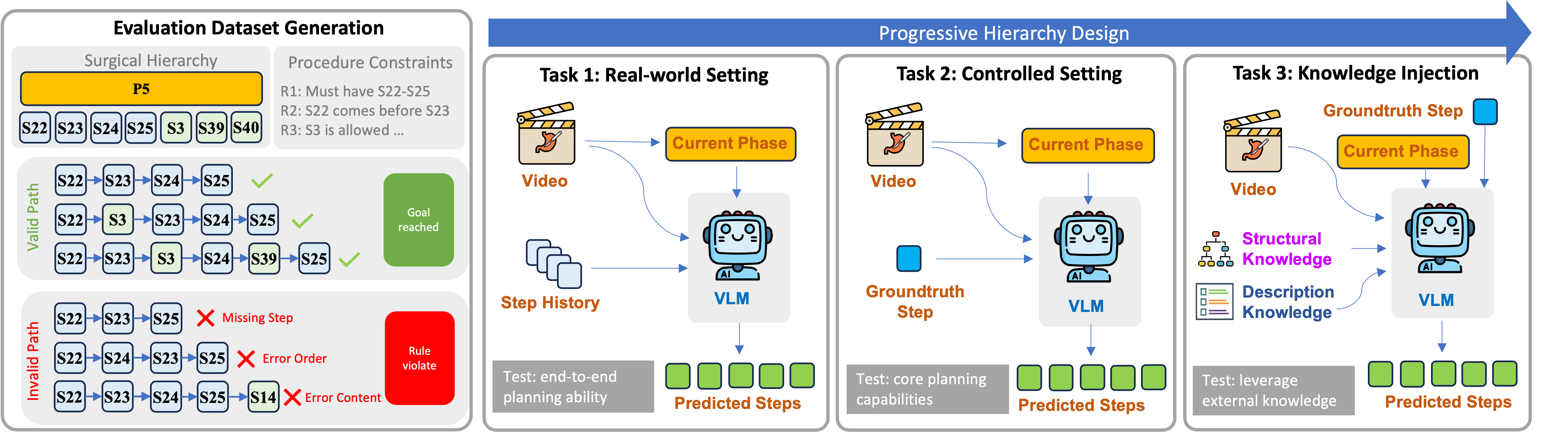}
    \caption{\textbf{Evaluation for Surgical Planning.} Left: Protocol adherence evaluation set construction via expert-defined protocol rules, separating valid and invalid step sequences. Right: Progressive evaluation of Video-LLMs, from end-to-end planning to planning with ground-truth steps and injected knowledge.}
    \label{fig:meta-eval}
\end{figure*}
\section{Can VLMs Generate Protocol-Adherent Surgical Plans?}

\subsection{Experimental Setup}
\textbf{Dataset.}
Experiments are conducted on MultiBypass140~\citep{Lavanchy2024MultiCentric}. To focus on logical planning rather than temporal boundary detection, we segment each video into 5,032 discrete step-level clips based on expert annotations. Each clip preserves its full temporal context (up to ten minutes), ensuring that the model has access to the complete visual evidence of the ongoing maneuver.

\textbf{Models.}
We evaluate DAMO (VideoLLaMA3-7B)~\citep{Zhang2025VideoLLaMA3}, LLaVA-NeXT-Video-7B~\citep{Li2024LLaVANextInterleave}, Qwen2.5/3-VL~\citep{Bai2025Qwen25VL,yang2025qwen3} (7B/32B), and medical models Hulu-Med~\citep{Jiang2025HuluMed} (7B/32B) and Lingshu~\citep{Xu2025LingShu} (7B/32B). All models are evaluated zero-shot (temperature = 0, max output = 2048 tokens) on an NVIDIA H200 GPU.

\subsection{Progressive Task Formulation}\label{task_define}

To disentangle visual perception from procedural reasoning, we define progressively constrained planning tasks. (Figure~\ref{fig:meta-eval})

\textbf{Real-World Setting: End-to-End Planning.}
\emph{Task 1}: Models are given raw surgical video clips, the current phase label, the history of completed steps, and a set of candidate step labels. They must infer the current procedural state from visual evidence and generate a plausible future step sequence. This setting mimics a real-world intraoperative assistant that must simultaneously ground its understanding in visual evidence (What is happening now?) and extrapolate future steps.

\textbf{Controlled Setting: Planning with Explicit State.}
\emph{Task 2}: To isolate planning from perception, models are additionally provided with the current step identity, while retaining video input. This removes ambiguity about the procedural state and allows us to directly assess the model’s ability to plan future steps from a fixed starting point.

\textbf{Knowledge Injection for Surgical Planning.}
\emph{Task 3}: We further investigate how external medical expertise modulates planning quality by injecting three forms of knowledge into the Task 2 setup:
\begin{enumerate}[label=3.\arabic*,topsep=2pt,itemsep=0pt,parsep=0pt]
  \item Structural knowledge: phase-step hierarchy.
  \item Semantic knowledge: expert-written natural language descriptions of the phases and steps.
  \item Combined knowledge: both knowledge.
\end{enumerate}
The structural, semantic, and combined knowledge prompts contain approximately 400, 900, and 1,300 tokens, respectively.

\subsection{Planning Output}
Each model produces a unified structured output following a fixed JSON schema:
\begin{itemize}[topsep=2pt,itemsep=0pt,parsep=0pt,leftmargin=*]
\item Remaining steps to complete the current phase
\item Next phase (as a phase name)
\item Reasonable step sequence for the next phase
\item Explanation (brief justification)
\end{itemize}
This output supports two planning tasks under a unified evaluation protocol: 

\textbf{{current-phase completion planning}}, evaluated on the combined plan of completed steps, the current step, and predicted remaining steps;

\textbf{Next-phase planning}, evaluated on the generated step sequence for the predicted next phase, isolates planning ability from perception noise.

\subsection{Evaluation}

We evaluate VLM surgical planning by \emph{protocol adherence accuracy}.
Each generated step sequence is checked against the expert-defined protocol rules introduced in Section~\ref{rules}. A sequence is labeled as valid if it satisfies all rules for the target phase, and invalid if it violates any safety-relevant rule. Accuracy is computed as the fraction of generated sequences labeled valid.

For Task~1, we additionally report \emph{current step recognition}, defined as exact matching between the predicted and ground-truth current step. This metric isolates errors in video-to-state grounding from errors in protocol-adherent planning. We do not evaluate exact next-phase prediction, because our goal is not to predict the next observed transition from the current video state. Instead, the next-phase planning directly evaluates whether the model can generate a protocol-adherence plan for the next phase.

\begin{table*}[t]
  \centering
  \small
  \setlength{\tabcolsep}{3pt}
  \resizebox{\textwidth}{!}{%
  \begin{tabular}{cccccccccccc}
    \toprule
    \textbf{Task} &
    \textbf{Metric} &
    \textbf{Hulu-7B} &
    \textbf{Hulu-32B} &
    \textbf{Qwen 2.5-7B} &
    \textbf{Qwen 2.5-32B} &
    \textbf{DAMO} &
    \textbf{Lingshu-7B} &
    \textbf{Lingshu-32B} &
    \textbf{LLaVA} &
    \textbf{Qwen 3-7B} &
    \textbf{Qwen 3-7B Think} \\
    \midrule
    \multirow{3}{*}{Real-world}
      & StepAcc    & 23.5\% & 34.0\% & 21.7\% & 31.4\% & 14.4\% & 26.1\% & \textbf{39.4\%} & 2.8\% & 28.3\% & 28.2\% \\
      & Current    & 10.3\% & 23.8\% & 2.9\%  & 20.9\% & 8.8\%  & 7.4\%  & \textbf{31.1\%} & 1.3\% & 21.1\% & 20.9\% \\
      & Next       & 2.9\%  & 45.6\% & 2.4\%  & 31.8\% & 1.4\%  & 1.0\%  & \textbf{46.5\%} & 0.1\% & 22.1\% & 22.1\% \\
    \midrule
    \multirow{2}{*}{Controlled}
      & Current    & 27.9\% & 40.2\% & 13.7\% & 48.6\% & 22.5\% & 22.3\% & \textbf{51.1\%} & 20.5\% & 47.3\% & 47.1\% \\
      & Next       & 2.7\%  & 31.0\% & 2.9\%  & 29.4\% & 1.5\%  & 4.5\%  & \textbf{40.7\%} & 0.1\%  & 26.1\% & 25.8\% \\
    \bottomrule
  \end{tabular}%
  }
  \caption{
  Protocol adherence accuracy (\%) across models.
 Task~1 (real-world) evaluates current step recognition (StepAcc) and downstream planning. Task~2 (controlled) focuses on phase-aware planning.
 Current/Next denote current-phase completion and next-phase planning.  Best results in bold. Qwen: Qwen-VL model series.
}
  \label{tab:task1_task2_results}
\end{table*}

\begin{table*}
  \centering
  \small
  \setlength{\tabcolsep}{3pt}
  \resizebox{\textwidth}{!}{%
  \begin{tabular}{cccccccccccc}
    \toprule
    \textbf{Task} &
    \textbf{Metric} &
    \textbf{Hulu-7B} &
    \textbf{Hulu-32B} &
    \textbf{Qwen 2.5-7B} &
    \textbf{Qwen 2.5-32B} &
    \textbf{DAMO} &
    \textbf{Lingshu-7B} &
    \textbf{Lingshu-32B} &
    \textbf{LLaVA} &
    \textbf{Qwen 3-7B} &
    \textbf{Qwen 3-7B Think} \\
    \midrule
    \multirow{2}{*}{Structural}
      & Current  & 42.8\% & 66.0\% & 63.0\% & \textbf{73.0\%} & 43.0\% & 46.1\% & 71.1\% & 23.4\% & 63.2\% & 63.1\% \\
      & Next     & 51.2\% & 49.2\% & 64.3\% & 68.7\% & 38.0\% & 62.6\% & \textbf{70.5\%} & 22.9\% & 64.5\% & 64.8\% \\
    \midrule
    \multirow{2}{*}{Description}
      & Current  & 26.8\% & 46.5\% & 14.0\% & 44.8\% & 26.9\% & 18.5\% & 49.7\% & 14.1\% & \textbf{53.2\%} & 53.1\% \\
      & Next     & 7.7\%  & 26.7\% & 11.5\% & \textbf{56.7\%} & 5.5\%  & 0.9\%  & 52.8\% & 0.1\%  & 54.4\% & 54.3\% \\
    \midrule
    \multirow{2}{*}{Combined}
      & Current  & 36.2\% & 67.6\% & 55.0\% & 69.1\% & 38.6\% & 34.6\% & \textbf{70.3\%} & 19.7\% & 61.6\% & 61.5\% \\
      & Next     & 41.7\% & 76.5\% & 60.8\% & \textbf{89.2\%} & 53.8\% & 56.5\% & 82.1\% & 21.3\% & 69.3\% & 69.0\% \\
    \bottomrule
  \end{tabular}%
  }
  \caption{
     Protocol adherence accuracy (\%) of Task~3 under different knowledge injection settings.  Current/Next denote current-phase completion and next-phase planning. Best results in bold. Qwen: Qwen-VL model series.
  }
  \label{tab:task3_combined_results}
\end{table*}

\subsection{Results and Analysis }
\subsubsection{Perception Bottleneck: Video Grounding Limits Planning}

Task~1 reflects the real-world setting where models must infer procedural state directly from video. As shown in Table~\ref{tab:task1_task2_results}, even the strongest model in our comparison (Lingshu-32B) achieves only 39.4\% step recognition accuracy, resulting in substantial downstream planning errors. A clinician-led analysis reveals two dominant perception failure modes.

\paragraph{Confusing exploration with procedural steps.}
Models often fail to distinguish long, repetitive exploratory video segments (e.g., tissue retraction to locate target organs or vessels) from well-defined surgical steps. 
For example, prolonged tissue exploration before definitive vessel exposure is often misclassified as a subsequent surgical step. As a result, models may prematurely conclude that a phase objective has been met, producing implausible plans that omit essential preparatory steps.

\paragraph{Failing to recognize repeated steps.}
Models also struggle when the same step appears multiple times within a phase.   During P2, steps such as horizontal stapling, retrogastric dissection, vertical stapling, and hemostasis may repeat multiple times. Models often misinterpret such repetition as phase progression or completion, leading to incorrect phase status estimation and subsequently flawed planning.

These perception errors propagate directly to planning. For most small models, protocol adherence accuracy remains below 25\%, while for larger models it is consistently lower for current-phase completion than for next-phase planning. This cascading failure indicates that step recognition errors directly undermine protocol-adherent planning. Importantly, this should not be interpreted as limited reasoning ability, as larger models perform well on next-phase planning. Rather, the results expose a dominant perception bottleneck: without reliable video-procedure alignment, even strong language backbones fail to produce protocol-adherent plans, indicating that end-to-end surgical planning is constrained primarily by video understanding rather than higher-level reasoning.

\subsubsection{Under-Constrained Reasoning: Semantically Plausible but Procedurally Invalid Plans}\label{Controlled Setting}
Task~2 removes perceptual ambiguity by providing the current step explicitly.  Although performance improves relative to Task~1, planning quickly saturates: current-phase completion remains below 52\%, and next-phase planning accuracy falls below 5\% for small models.

\paragraph{Semantically plausible but procedurally invalid plans.}
Without explicit procedural guidance, models default to semantic proximity rather than procedural logic. As a result, generated plans often omit critical steps or assemble loosely related steps that violate protocol constraints for the phase.
For example, sequences may mix steps from the omentum division phase (e.g., omentum exposure and omental transection) with unrelated steps from gastrojejunal anastomosis, including biliary limb measurement or jejunal opening.

This indicates that, although Video-LLMs encode general surgical knowledge, particularly in medical VLMs such as Lingshu and Hulu-Med, their planning space remains under-constrained. Without explicitly encoded procedural constraints, models fail to produce protocol-adherent step-level plans, leading to performance saturation even when perceptual uncertainty is removed.

\subsubsection{Knowledge Injection: Medical Guidance for Planning}
Task~3 introduces explicit medical knowledge to constrain planning, revealing how different forms of knowledge affect model behavior.

\paragraph{Structural Knowledge constrains planning effectively.}
For most models, structural knowledge (Task~3.1) is the most effective intervention. By explicitly specifying the phase-step hierarchy, procedural structure sharply narrows the space of admissible plans. This leads to large and consistent gains over Task~2 in both current-phase completion and next-phase planning (Table~\ref{tab:task3_combined_results}), particularly for 7B-scale models. Generated plans exhibit fewer cross-phase intrusions and more reliably satisfy phase-level requirements under protocol adherence evaluation.
For example, when planning the gastrojejunal anastomosis phase, unrelated steps such as Petersen space exposure or biliary limb opening are more consistently excluded, which were frequently misincorporated in earlier tasks. 

Similarly, during jejunojejunal anastomosis planning, extraneous steps such as mesenteric defect exposure or mesenteric defect closure are largely eliminated.
These results show that concise structural constraints keep model generation within the protocol-defined scope of the target phase.

\paragraph{Semantic descriptions alone cannot enforce procedural correctness.}
In contrast, semantic descriptions without explicit structure (Task~3.2), as well as the combined setting (Task~3.3), underperform structural guidance alone for 7B-scale models, especially for next-phase planning(Table~\ref{tab:task3_combined_results}). Although models often express correct surgical intent, they frequently omit critical execution steps needed to complete the phase goal. Rich semantic prompts tend to promote narrative plausibility rather than enforce discrete procedural requirements.

For instance, after Petersen space closure, models may correctly identify jejunojejunal anastomosis as the next phase, but still fail to include essential steps such as biliary limb opening or alimentary limb measurement. Under long-context prompts, semantic information is often diluted, leading to shallow or generic plans. Models may also lose global procedural coherence, treating jejunojejunal anastomosis as the final phase and appending steps from later phases (e.g., mesenteric defect closure or cleaning and coagulation) in an unstructured manner. 
These failures show that semantic guidance alone insufficiently constrains the planning space.

\subsubsection{Model Capacity Modulates Knowledge Integration.}
Across all settings, larger models outperform their 7B-scale counterparts, reflecting stronger reasoning capacity and more reliable use of long-context inputs. Structural knowledge (Task~3.1) provides a stable benefit across model sizes, consistently improving planning performance relative to Task~2.

\paragraph{Small models struggle to combine multiple guidance.}
For 7B-scale models, combining semantic and structural knowledge (Task~3.3) consistently performs worse than structural constraints alone, particularly for next-phase planning (Table~\ref{tab:task3_combined_results}). This suggests that limited-capacity models have difficulty integrating heterogeneous information sources, leading to weaker adherence to procedural constraints.

\paragraph{Larger models better exploit combined knowledge.}
For 32B-scale models, Task~3.3 achieves the best next-phase planning performance (Table~\ref{tab:task3_combined_results}). This indicates that sufficient capacity enables models to leverage semantic descriptions to refine intent-level reasoning while still relying on structural constraints to maintain procedural validity. Notably, structural knowledge alone remains highly competitive even at 32B-scale, suggesting that explicit procedural structure is a robust aid for protocol-adherent surgical planning across model capacities.

Overall, these results suggest that model capacity governs the ability to integrate multiple forms of knowledge. Semantic enrichment becomes beneficial primarily when sufficient capacity is available to reconcile heterogeneous guidance without compromising procedural validity.

\section{Conclusion}
This work shows that prevailing evaluations of surgical planning conflate two distinct concepts: task completion and protocol adherence. Trajectory similarity, used as a proxy for completion in benchmarks such as Ego4D, is already a poor approximation of that target. More fundamentally, completion is not the right goal in surgery: what matters is whether a plan adheres to the safety constraints defined by clinical protocols, which exist independently of any observed execution. We introduce an evaluation framework grounded in expert-defined protocol rules that directly measures this. Under this framework, widely used sequence similarity metrics fail systematically: they reject most protocol-adherent plans while accepting protocol-violating ones. LLM-based judges, despite strong semantic understanding, frequently fail to enforce strict procedural dependencies. Through progressive evaluation of Video-LLMs, we identify two separable failure modes: unreliable video-to-state grounding and under-constrained reasoning over the space of valid plans. Structural knowledge injection provides the most consistent gains across model scales, whereas semantic descriptions alone are insufficient and their combination benefits only larger models. Together, these findings establish protocol adherence as the appropriate evaluation standard for surgical planning and identify the technical gaps that must be closed before VLMs can be deployed safely in this domain.

\section*{Limitations}

This study has several limitations that suggest directions for future work.

First, our expert-defined protocol constraints are manually constructed from surgical principles and dataset-specific annotation protocols. This design provides interpretability and clinical grounding, but does not readily scale to new procedures, institutions, or surgical domains. Developing automated or semi-automated methods for constructing such protocol rules, potentially with LLMs assisting clinicians in formalizing surgical knowledge, remains an important challenge.

Second, our evaluation treats protocol adherence as a binary criterion: a plan either satisfies all rules or it does not. Although appropriate for a safety floor, this formulation does not capture finer-grained aspects of planning quality, such as efficiency, redundancy, or preferences among multiple valid procedural paths. Future work could explore more nuanced, adherence-aware metrics that go beyond a pass/fail safety check without reverting to single-path assumptions.

Third, our experiments are conducted on a limited set of surgical datasets with rich hierarchical annotations. At present, few publicly available datasets provide phase-step hierarchies with sufficient granularity and consistency to support protocol-adherence evaluation. Extending this framework to additional procedures will require broader annotation efforts or alternative forms of weak or implicit supervision.

\bibliography{custom}

\appendix

\section{Experiment details}\label{appendix}

\begin{table*}[t]
\centering
\small
\caption{Surgical step abbreviations used in Figure~\ref{fig:comparison}.}
\label{tab:step_names}
\resizebox{\textwidth}{!}{
\begin{tabular}{ll ll ll}
\toprule
\textbf{ID} & \textbf{Step name} &
\textbf{ID} & \textbf{Step name} &
\textbf{ID} & \textbf{Step name} \\
\midrule
S1  & cavity\_exploration              & S16 & biliary\_limb\_measurement        & S31 & alimentary\_limb\_opening \\
S2  & trocar\_placement                & S17 & jejunum\_opening                  & S32 & jejunojejunal\_stapling \\
S3  & retractor\_placement             & S18 & gastrojejunal\_stapling           & S33 & jejunojejunal\_defect\_closing \\
S4  & crura\_dissection                & S19 & gastrojejunal\_defect\_closing    & S34 & jejunojejunal\_anastomosis\_reinforcement \\
S5  & his\_angle\_dissection           & S20 & mesenteric\_opening               & S35 & staple\_line\_reinforcement \\
S6  & horizontal\_stapling             & S21 & jejunal\_section                  & S36 & mesenteric\_defect\_exposure \\
S7  & retrogastric\_dissection         & S22 & gastric\_tube\_placement          & S37 & mesenteric\_defect\_closing \\
S8  & vertical\_stapling               & S23 & jejunal\_clamping                 & S38 & anastomosis\_fixation \\
S9  & gastric\_remnant\_reinforcement  & S24 & ink\_injection                    & S39 & coagulation \\
S10 & gastric\_pouch\_reinforcement    & S25 & visual\_assessment                & S40 & irrigation\_aspiration \\
S11 & gastric\_opening                 & S26 & gastrojejunal\_anastomosis\_reinforcement & S41 & parietal\_closure \\
S12 & omental\_lifting                 & S27 & petersen\_space\_exposure         & S42 & trocar\_removal \\
S13 & omental\_section                 & S28 & petersen\_space\_closing          & S43 & calibration \\
S14 & adhesiolysis                     & S29 & biliary\_limb\_opening            & S44 & drainage\_insertion \\
S15 & treitz\_angle\_identification    & S30 & alimentary\_limb\_measurement     & S45 & specimen\_retrieval \\
\bottomrule
\end{tabular}
}
\end{table*}

\begin{table*}[t]
  \centering
  \small
  \caption{Surgical phase abbreviations used in the study.}
  \label{tab:phase_names}
  \begin{tabular}{ll ll}
    \toprule
    \textbf{ID} & \textbf{Phase name} &
    \textbf{ID} & \textbf{Phase name} \\
    \midrule
    P1 & Preparation               & P7  & Petersen space closure \\
    P2 & Gastric pouch creation    & P8  & Jejunojejunal anastomosis \\
    P3 & Omentum division          & P9  & Mesenteric defect closure \\
    P4 & Gastrojejunal anastomosis & P10 & Cleaning \& coagulation \\
    P5 & Anastomosis test          & P11 & Disassembling \\
    P6 & Jejunal separation        &     & \\
    \bottomrule
  \end{tabular}
\end{table*}

\begin{figure*}[]
    \centerline{\includegraphics[width=0.8\textwidth]{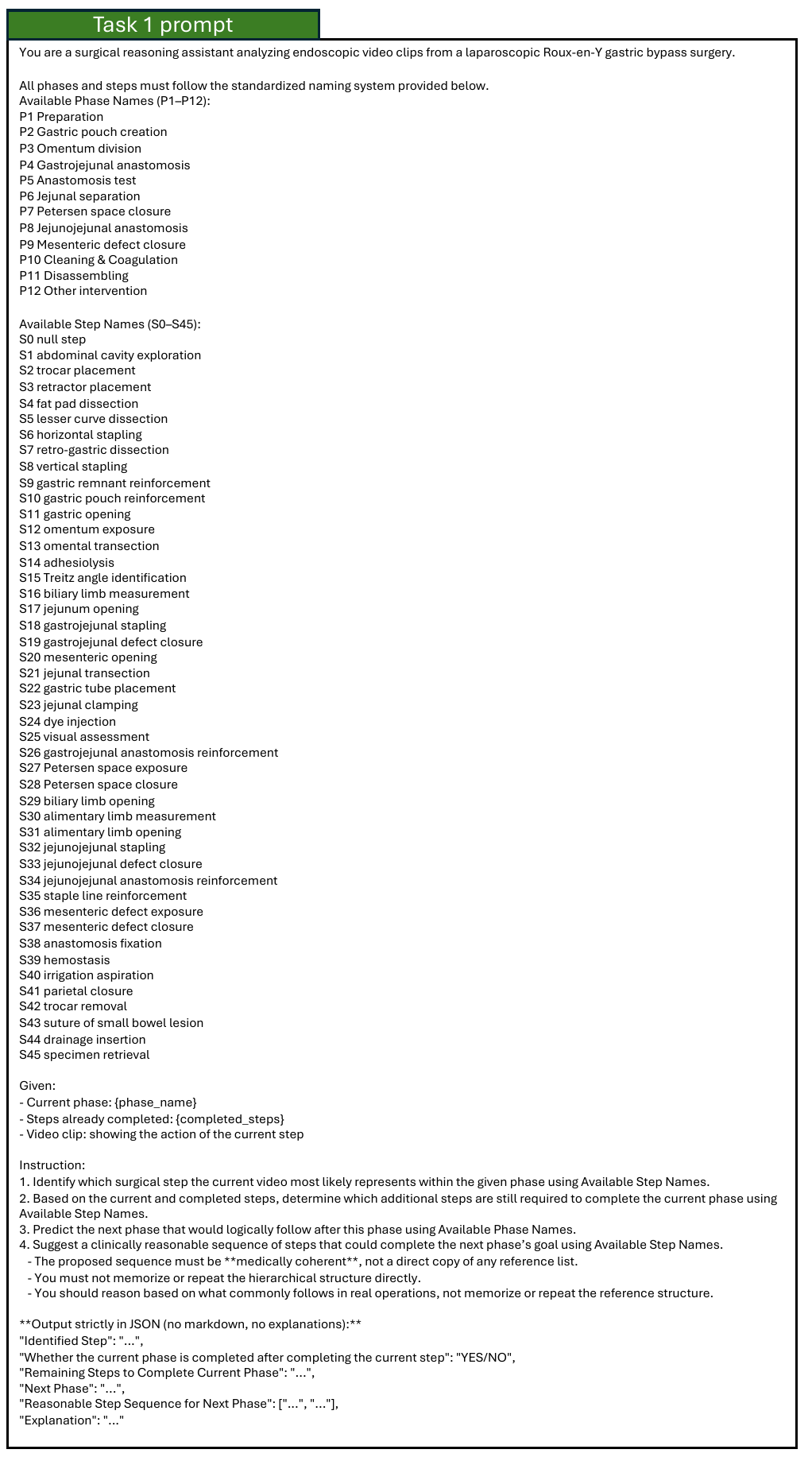}}
\caption{Task 1 prompt.}
\label{fig:task1_prompt}
\end{figure*}

\begin{figure*}[]
    \centerline{\includegraphics[width=0.8\textwidth]{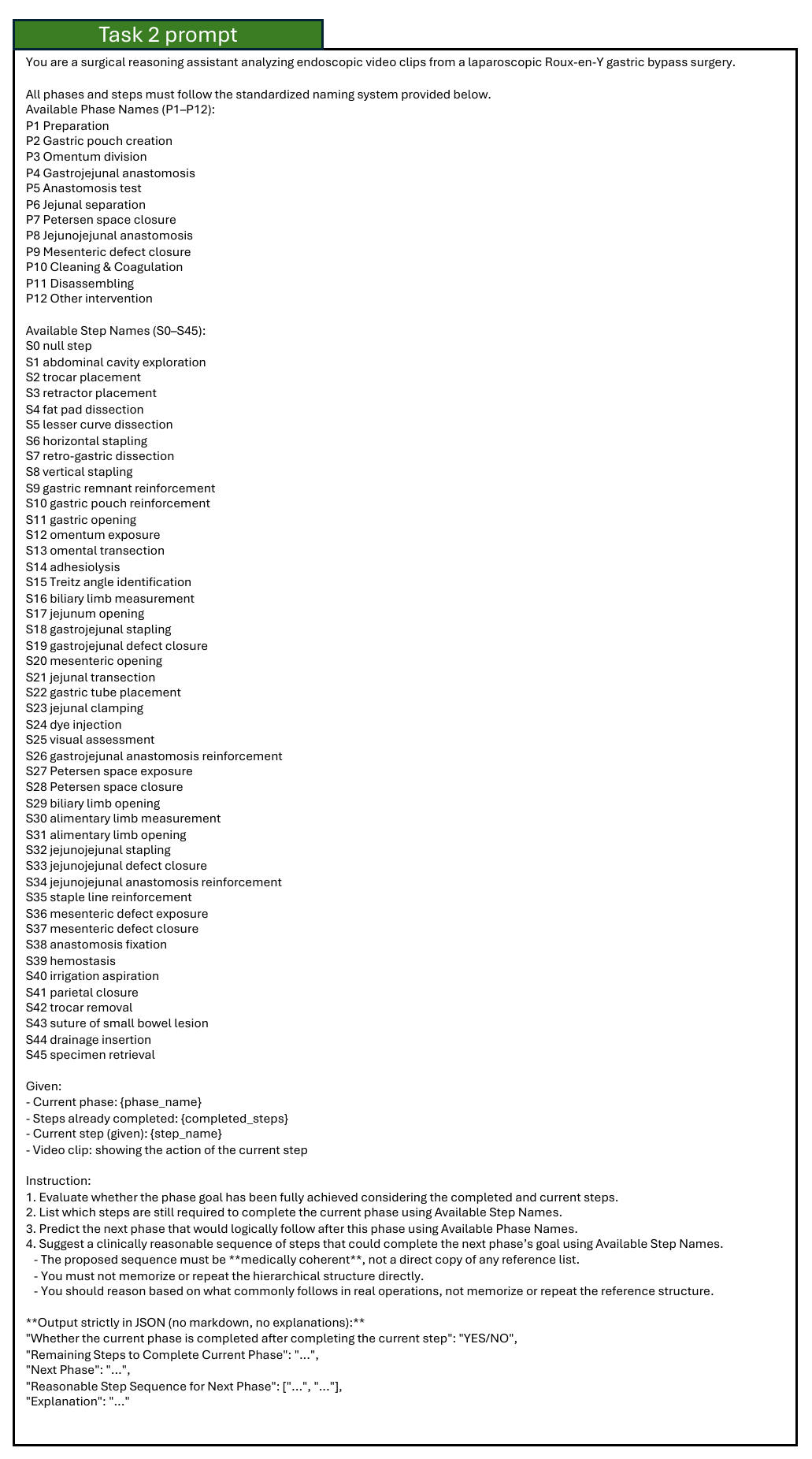}}
\caption{Task 2 prompt.}
\label{fig:task2_prompt}
\end{figure*}

\begin{figure*}[]
    \centerline{\includegraphics[width=0.8\textwidth]{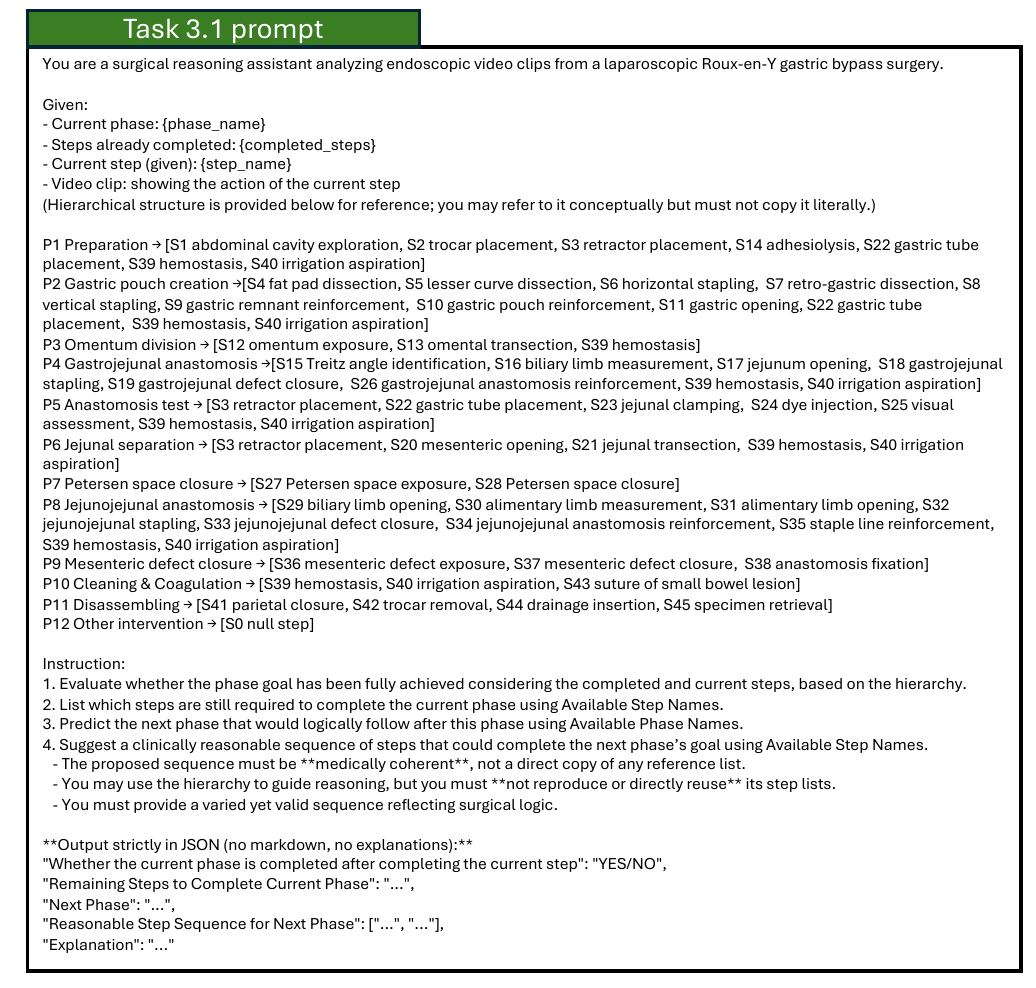}}
\caption{Task 3.1 prompt.}
\label{fig:task31_prompt}
\end{figure*}

\begin{figure*}[]
    \centerline{\includegraphics[width=0.8\textwidth]{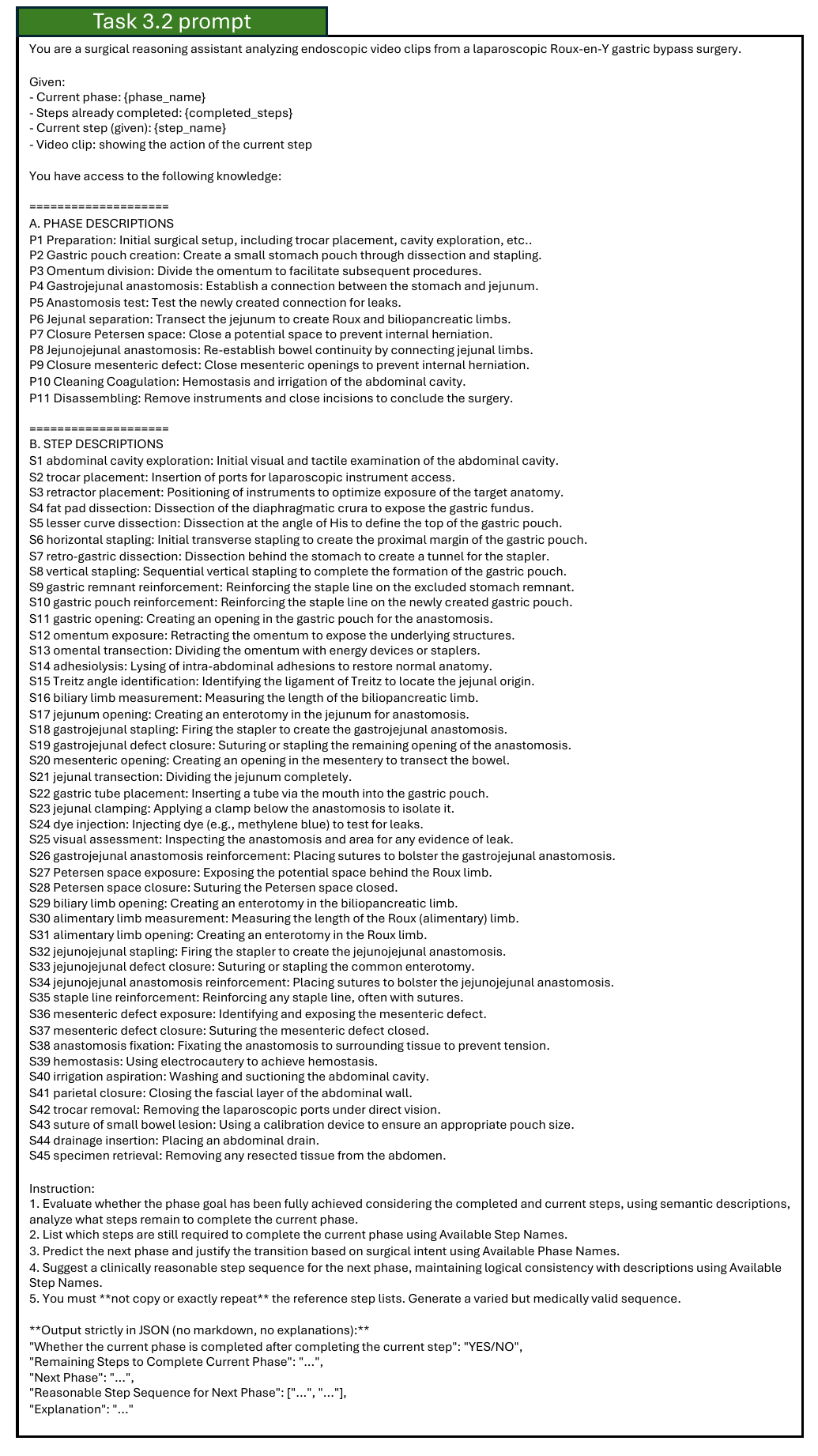}}
\caption{Task 3.2 prompt.}
\label{fig:task32_prompt}
\end{figure*}

\begin{figure*}[]
    \centerline{\includegraphics[width=0.6\textwidth]{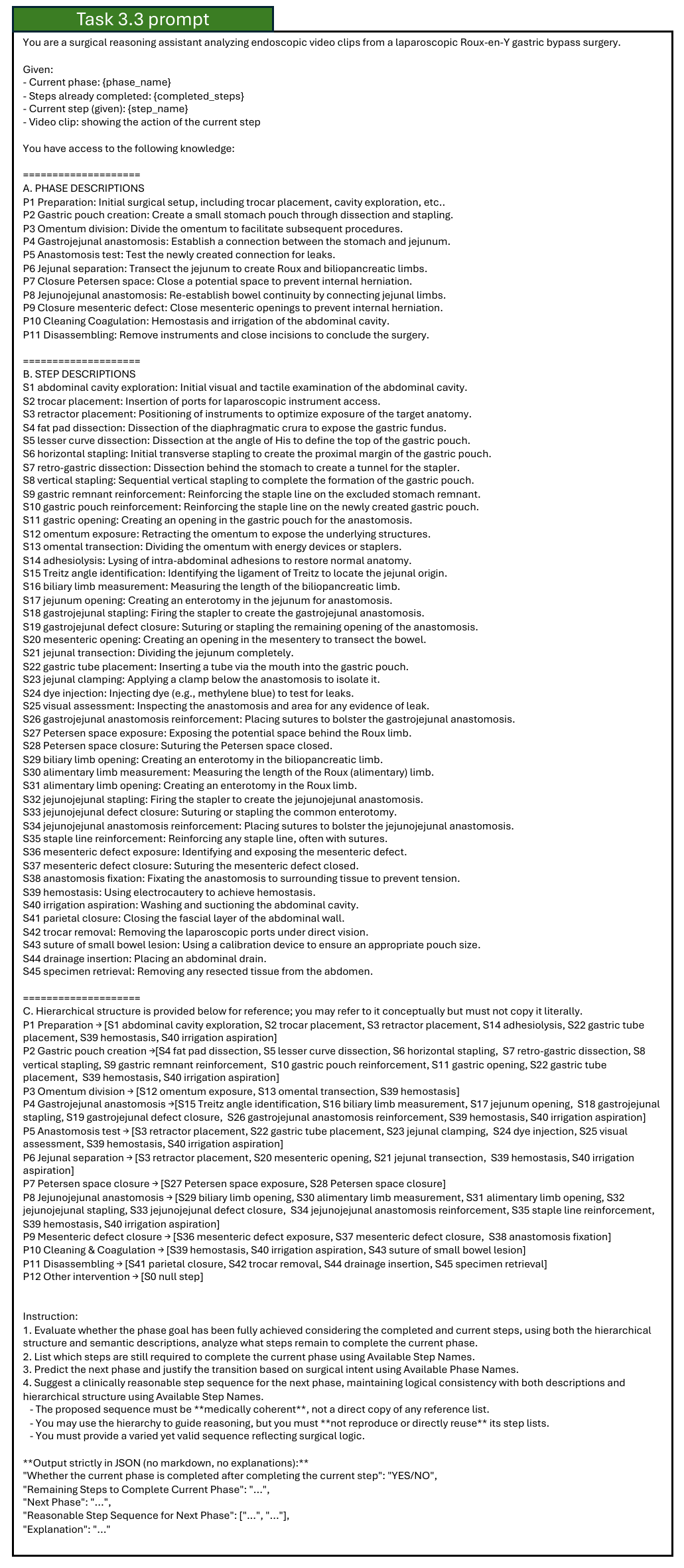}}
\caption{Task 3.3 prompt.}
\label{fig:task33_prompt}
\end{figure*}

\begin{figure*}[]
    \centerline{\includegraphics[width=0.5\textwidth]{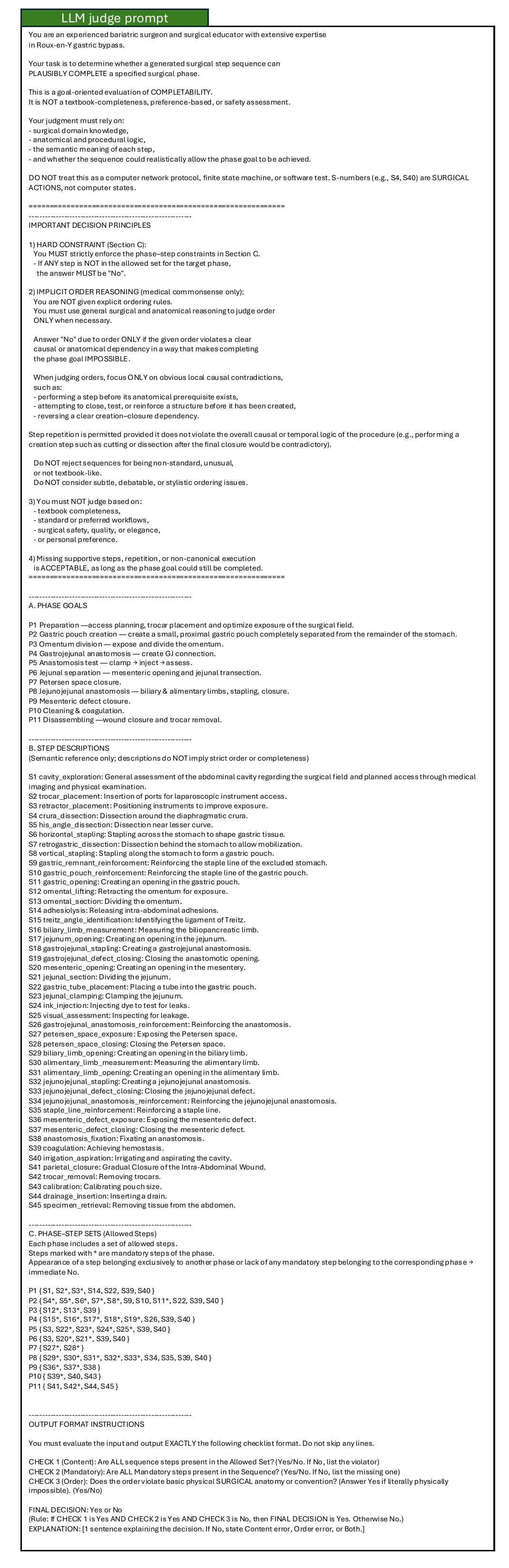}}
\caption{LLM judge prompt.}
\label{fig:llm_judge_prompt}
\end{figure*}

\end{document}